\documentclass[english]{maciarticle}

\usepackage{amsmath,amsbsy,amscd,amssymb,graphicx, epsfig,color,array}
\usepackage{float}
\usepackage{verbatim}
\usepackage{hyperref}
\usepackage{siunitx}

\newcommand{\PreserveBackslash}[1]{\let\temp=\\#1\let\\=\temp}
\newcolumntype{C}[1]{>{\PreserveBackslash\centering}p{#1}}
\newcolumntype{R}[1]{>{\PreserveBackslash\raggedleft}p{#1}}
\newcolumntype{L}[1]{>{\PreserveBackslash\raggedright}p{#1}}

\begin{document}

\title{Optimal market making by reinforcement learning}

\renewcommand{\thefootnote}{\arabic{footnote}}

\author[,$\flat$,$\dag$]{Matías Selser \footnote[1]{These two authors contributed equally}}

\author [,$\flat$,$\ddag$]{Javier Kreiner $^*$}
\author[$\dag$,$\sharp$]{Manuel Maurette}
\affil[$\flat$]{Almafintech S.A., Argentina, \textcolor{blue} { \{matias,javier\}@almafintech.com.ar} }
%
\affil[$\dag$]{Universidad Torcuato Di Tella, Buenos Aires, Argentina}
\affil[$\sharp$]{UCEMA, Buenos Aires, Argentina; Qontigo, a Deutsche Boerse company}
\affil[$\ddag$]{Universidad de Buenos Aires, Argentina}
\maketitle

\begin{abstract} We apply Reinforcement Learning algorithms to solve the classic quantitative finance Market Making problem, in which an agent provides liquidity to the market by placing buy and sell orders while maximizing a utility function. The optimal agent has to find a delicate balance between the price risk of her inventory and the profits obtained by capturing the bid-ask spread. We design an environment with a reward function that determines an order relation between policies equivalent to the original utility function. When comparing our agents with the optimal solution and a benchmark symmetric agent, we find that the Deep Q-Learning algorithm manages to recover the optimal agent.

\end{abstract}
\begin{mathsubclass}
68T07 - 91G15 - 91G30 - 91G60
\end{mathsubclass}

{\thispagestyle{empty}} 

\section{Introduction}

A classic problem in quantitative finance is Market Making (MM). A Market Maker's task is to provide liquidity to other participants by continuously placing buy and sell orders, whilst remaining profitable. The Market Maker usually has a number of advantages with respect to other traders such as lower transaction costs, the ability to send orders at a higher frequency without penalties, or even monetary rewards for the provision of liquidity. These advantages compensate the obligation to provide liquidity for a significant proportion of trading session time, with maximum limits to the quoted bid-ask spread and/or a minimum amount of visible liquidity that needs to be provided.

A possible formulation of the MM problem is to model the environment as a stochastic optimal control problem with a suitable utility function to maximize as objective, as in \cite{Avellaneda}. This setup abstracts from many of the complexities of the micro-structure of a real multi-agent exchange, but still addresses the fundamental trade-off between holding inventory risk and profiting by capturing the bid-ask spread through MM activity.

Reinforcement learning (RL) is a general framework that allows agents to learn optimal behaviours through interaction with their environment. In recent years it has been successfully applied to solve difficult challenges that had until then defied other methods. RL agents have been trained to play Atari games \cite{rl_atari}, play Go \cite{rl_go}, manipulate robotic actuators \cite{rl_actuator}, and drive autonomous vehicles \cite{rl_driving}, among other applications, and many of these solutions display superhuman skills. It has proven particularly effective in situations where the environment can be simulated and the agent has access to virtually unlimited amounts of interaction data. The successes in other areas have spurred a stream of research that tries to apply similar techniques in the financial domain. Some applications of RL in finance include trading bots \cite{rl_trading_bot}, risk optimization \cite{rl_risk}, and portfolio management \cite{rl_portfolio}.  

In this paper we show how the MM problem can be reformulated and the optimal solution recovered by training an agent using RL methods. 

\section{Model and Algorithm}
We consider an asset whose price $s$ changes according to:

\vspace{-0.2cm}
\begin{equation*}
    ds_t = \sigma dW_t,
\end{equation*}

\vspace{-0.1cm}

\noindent where $W_t$ is a one-dimensional Brownian motion and $\sigma$ is constant. The Market Maker can control the prices $p^b_t$ and $p^a_t$ at which she offers to buy or sell the security, respectively. The buy or sell orders will be 'hit' or 'lifted' by Poisson processes with rates $\lambda(\delta^b_t$), $\lambda(\delta^a_t)$, that depend on the distance between the bid and ask prices, and the asset price: $\delta^b_t=s_t-p^b_t$, $\delta^a_t=p^a_t-s_t$, respectively. We let the rate function be $\lambda(\delta) = A \exp(-k\delta)$, a decreasing function of $\delta$, again following \cite{Avellaneda}. 

The cash $X$ of the agent evolves according to the following equation:

\vspace{-0.2cm}
\begin{equation*}
    dX_t = p^a_t dN^a_t -  p^b_t dN^b_t.
\end{equation*}

\vspace{-0.1cm}

That is, it accumulates whenever the asset is sold and decreases when it is bought, where $N^a_t$, $N^b_t$ are Poisson processes with rates $\lambda(\delta^b_t$), $\lambda(\delta^a_t)$. The inventory of the asset held at time $t$, denoted $q_t$, follows the dynamics:

\vspace{-0.2cm}

\begin{equation*}
    dq_t =  dN^b_t - dN^a_t.
\end{equation*}

\vspace{-0.1cm}

The agent's objective is to maximize the expected value of a concave utility function of the final wealth:

\vspace{-0.2cm}

\begin{equation}
\max_{\{p^b_t\}, \{p^a_t\}} \mathbb{E} [-\exp(-\beta (X_T + q_T s_T))],
\label{eq:objective_function}
\end{equation}

\vspace{-0.1cm}

\noindent where $w_T = X_T + q_T s_T$ is the agent's wealth at time $T$ (cash plus liquidation value of the inventory). The utility function adds risk-aversion to the agent's preferences, as opposed to pure maximum profit seeking. A first order approximation of the optimal agent was obtained analytically in \cite{Avellaneda}, resulting in a closed form solution:  $p^{b}_{t} := \rho - \frac{\Phi_t}{2}$ for the bid price and $p^{a}_{t}:= \rho + \frac{\Phi_t}{2}$ for the ask price, where $\Phi_t := \frac{2}{\beta} \ln(1 + \frac{\beta}{k})$ is the bid-ask spread, and $\rho := s_t -\beta \sigma^2 (T-t)  q_t$ is the mid-price adjusted by a factor times the negative inventory.

In the present setup this is equivalent to the mean-variance formulation, where the objective to maximize is replaced by $\max_{\{p^b_t\}, \{p^a_t\}} \mathbb{E} [w_T]-\frac{\kappa}{2} \mathbb{V}[w_T]$ (for a suitable $\kappa$). Assuming independence of changes in wealth across time-periods we obtain  (see \cite{Ritter}): 

\vspace{-0.2cm}

\begin{equation}
\mathbb{E} [w_T]-\frac{\kappa}{2} \mathbb{V}[w_T] = \sum \mathbb{E} [\delta w_t]-\frac{\kappa}{2} \sum \mathbb{V}[\delta w_t], 
\end{equation}

\vspace{-0.1cm}

We can now proceed to design the reward that our RL agent will receive at each time step: $\delta w_t - \frac{\kappa}{2} (\delta w_t - \hat{\mu})^2 $, where $\hat{\mu}$ is a running estimate of the mean of the single period returns $\delta w_t$.

\begin{figure}[h!]
    \centering
    \includegraphics[width=0.5\textwidth]{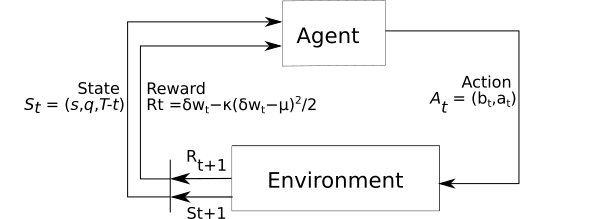}
    \caption{Reinforcement learning problem structure}
    \label{fig:reinfocement_struct}
\end{figure}

Our agent interacts with her environment in the traditional RL setting, as illustrated in Figure \ref{fig:reinfocement_struct}. To this purpose we discretize time at a reasonable resolution $dt$ and at each step provide the agent with an observation of the state of the world $S_t$. Here, we let $S_t = (s_t, q_t, T-t)$, a tuple with the asset mid-price $s_t$, the current inventory $q_t$, and the time remaining until the end of the period $T-t$. Based on said state the agent chooses the action $A_t = (b_t, a_t)$, which is a tuple of two numbers, the bid price and ask price she wants to quote. Come the next step the agent will receive a stochastic reward $R_{t+1} = \delta w_t - \frac{\kappa}{2} (\delta w_t - \hat{\mu})^2$ and the new state of the world $S_{t+1}$.

The cumulative discounted reward is defined as $G_t := \sum_{s=t}^{s=T} \gamma^{s-t}R_{s+1}$, and the agent's objective is to maximize its expectation. Setting $\gamma = 1.0$ and under the assumptions above, this is approximately equivalent to maximizing the objective function in Equation \ref{eq:objective_function}.

\vspace{-0.2cm}

\vspace{-0.6cm}

\section{Methods and Experiments}

We train agents with two different RL algorithms, Deep Q-Learning (DQN) and Tabular Q-Learning (TQL), and compare the resulting agents with two benchmarks: the optimal agent and the symmetric agent, which quotes symmetrically around the mid-price with the same spread as the optimal agent. Both RL algorithms work by having an estimate $\hat{q}(s,a)$ of the state-action value function, defined as the expected value of the cumulative discounted reward given that we follow a policy $\pi$: $q_{\pi}(s,a) = \mathbb{E}_{\pi}[G_t]$. The estimate $\hat{q}(s,a)$ is updated at each interaction with the environment, continuously and simultaneously improving the policy and the estimate by using the update formula: 

\vspace{-0.2cm}
\begin{equation}
\hat{q}_{new}(s_t,a_t) \gets \hat{q}(s_t,a_t) + \alpha (r_t + \gamma  \max_a \hat{q}(s_{t+1},a) - Q(s_t,a_t)),
\label{eq:qlearning_update}
\end{equation}

\vspace{-0.1cm}

\noindent where in the case of TQL the estimate $\hat{q}(s,t)$ is stored as a table, and in DQN it is represented as a neural network and each update is used as an input-output pair to train the network through gradient descent. For both DQN and TQL we discretize the action space to have $n_a$ possible actions as in \cite{Spooner}, where the discrete actions represent the number of steps of size $d_a$ away from the mid-price. For the spread we use, as before, the optimal spread. From these two values the environment calculates the bid and ask prices quoted by the agent. In the case of TQL we also discretize each component of the state observation $S_t = (s_t, q_t, T-t)$, where $s_t$ is expressed as the number of steps $ds=\sigma \sqrt dt$ away from the starting price $s_0$, the inventory $q_t$ is in units of the asset, and $T-t$ is expressed as the number of time steps remaining until the end of the episode. Hence the components of the state are integers, the first two bounded in absolute value by the episode length, and the last is in the range $[0,T/dt]$. The number of possible states for TQL is then $(2\frac{T}{dt}+1)^2(\frac{T}{dt}+1)$.

We chose the following parameters for our simulation: $s_0 = 100$ (the starting price of the asset), $T = 1$ (the length of the simulation), $\sigma = 2.0$ (the standard deviation of the asset's price), $dt = 0.05$ (the discrete time increments), $q_0 = 0$ (the starting inventory), $\beta = 0.5$ (the agent's risk aversion), $\gamma = 1.0$ (the reward discount factor), $\kappa = 1.0$, $A = 137.45$, $k = 1.5$, $n_a = 21$ and $d_a = 0.2$ (the probability of 'lifting' or 'hitting' a price farther than $\frac{n_a-1}{2} d_a = 2.0$ away from the mid-price is very close to $0$ for $\lambda(\delta)$ given the chosen parameters, so we restrict quoted prices to the interval $[s_t-2.0,s_t+2.0]$).

Both Q-learning agents were trained using a learning rate $\alpha = 0.6$. DQN was trained for $\num[group-separator={,}]{1000}$ episodes, while TQL was trained for $\num[group-separator={,}]{5000000}$. For the DQN agent we used a network with two hidden fully connected layers of size 10 and ReLU activation functions, the network had a total parameter count of 381.

\section{Results}

After training the RL agents, we run 1000 simulations to evaluate their performance\footnote{Code for the simulations can be found at: \href{https://github.com/mselser95/optimal-market-making}{https://github.com/mselser95/optimal-market-making}}.

\vspace{-0.1cm}

\begin{figure}[!ht]
    \centering
    \begin{minipage}[b]{0.45\textwidth}
        \includegraphics[width=\textwidth]{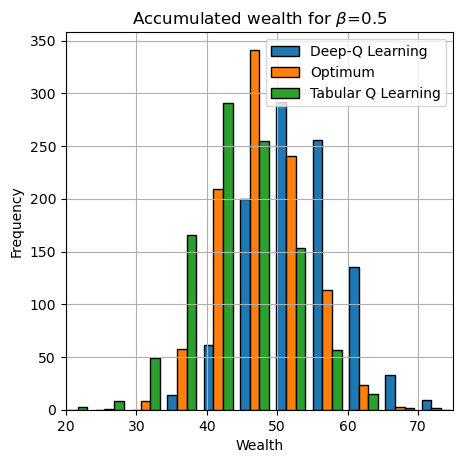}
        \caption{Wealth for $\beta$ = 0.5}
        \label{fig:comparison_hist_0.1}
    \end{minipage}
    \begin{minipage}[b]{0.45\textwidth}
        \includegraphics[width=\textwidth]{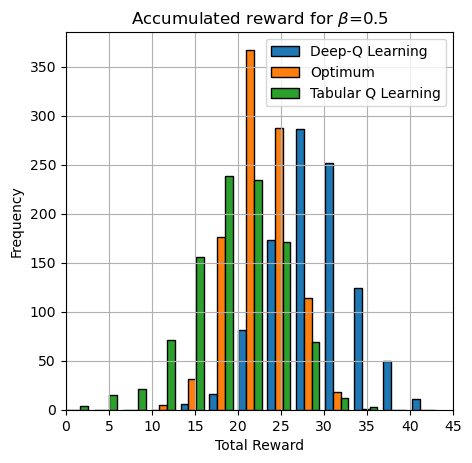}
    \caption{Accumulated reward for $\beta = 0.5$}
        \label{fig:comparison_hist_1}
    \end{minipage}
    \hfill
\end{figure}

\vspace{-0.2cm}

\begin{table}[!ht]
    \centering
    \begin{tabular}{| L{3.26cm} | C{2.75cm}| C{2.75cm} | C{2.75cm} | C{2.75cm} | }
     \hline
                    &   Optimal Agent     &   Symmetric   &   Tabular Q       &   Deep-Q      \\ \hline
    Mean Wealth     &   47.79       &   57.67       &   44.21           &   53.47       \\ \hline
    Std. Dev. Wealth&   6.09       &   11.86       &   7.08           &   6.67       \\ \hline
    Sharpe Ratio    &   7.83       &   4.86       &   6.24           &   8.00       \\ \hline
    Mean Cum. Reward     &   22.46       &   -7.17       &   19.19           &   29.04       \\ \hline
    Utility Estimate&   -2.63e-9       &   -4.34e-6       &   -1.49e-6           &   -2.20e-10       \\ \hline
    \end{tabular}
    \caption{1000 simulations for $\beta = 0.5$}
    \label{tab:table_strat_tr}
\end{table}

\vspace{-0.2cm}

To compare the different methods and benchmarks, we calculate the mean and standard deviation of final wealth, and its Sharpe ratio. We also obtain the mean cumulative reward (what our RL agents try to maximize) and the Monte Carlo estimate of the original utility function in Equation \ref{eq:objective_function}. 
The Symmetric agent obtains the highest mean wealth, but at the cost of high dispersion around that value. Surprisingly, the DQN agent manages to outperform the agent obtained in \cite{Avellaneda}, the reason could be that said agent uses a first order approximation of the order arrival rate. In contrast, the TQL agent does not manage to achieve the same level of performance, despite having been trained for several orders of magnitude more episodes. We believe that the cause is the huge size of the table holding the estimate of the value function, which potentially could have as many as $(2\frac{T}{dt}+1)^2(\frac{T}{dt}+1) n_a = \num[group-separator={,}]{675364200}$ entries, although many of the states corresponding to these entries are extremely unlikely or impossible to reach. In practice, our agent had $\num[group-separator={,}]{1925393}$ non-zero table entries after training, and we would need to train for longer to obtain a more precise estimate of the optimal policy's q-function and thus a better performing agent. In conclusion, we found that Deep Q-Learning, which shares weights across states and actions, can produce a more austere approximation than Tabular Q-Learning and requires far less experience to train effectively.  
\vspace{-0.2cm}

\section{Discussion and Future Work}

During our experiments we found that the training of RL agents is finicky and very sensitive to hyper-parameter settings. Finding a solution to which the algorithms converge consistently, regardless of random seed (used for parameter initialization and for generating the episodes) is non-trivial. Effectively, this lack of robustness, combined with the difficulty in explaining the inner workings of machine learning models, remains a significant obstacle to their use in production. We were pleasantly surprised, however, that the solution found by DQN was superior to the first order approximation derived in \cite{Avellaneda}. We are hopeful that in future, with superior tools to improve convergence and better error measures, these techniques will become part of the standard Quant toolbox.

As future research we would like to characterize the final wealth distribution, so as to be able to choose the $\kappa$ parameter in a principled manner. We would also like to obtain the optimal agent (without linearly approximating the order arrival rate function) through other methods and compare it to the solution obtained here. Additionally, we would like to use similar techniques in more complex settings, where we take into account the order book micro-structure of the market or we simulate a multi-agent environment where agents with diverse behaviours interact.
\vspace{-0.2cm}
\section*{Acknowledgments}
We thank Sebastian Di Tella and Lisandro Kaunitz for useful comments and for proofreading the article.
\vspace{-0.6cm}

\end{document}